\documentclass[letterpaper, 10 pt, conference]{ieeeconf}  

\IEEEoverridecommandlockouts                              

\usepackage{graphicx}

\title{\LARGE \bf
Diffusing in Someone Else's Shoes: \\
Robotic Perspective-Taking with Diffusion
}

\author{Josua Spisak$^{1}$, Matthias Kerzel$^{1}$, Stefan Wermter$^{1}$
\thanks{*The authors gratefully acknowledge support from the DFG (CML, MoReSpace,
LeCAREbot), BMWK (VERIKAS), and the European Commission
(TRAIL, TERAIS).}
\thanks{$^{1}$Josua Spisak, Matthias Kerzel and Stefan Wermter are with the Knowledge Technology (WTM) group, Department of Informatics,
        University of Hamburg, 22527 Hamburg, Germany.
        {\tt\small josua.spisak@uni-hamburg.de}}}%

\begin{document}

\maketitle
\thispagestyle{empty}
\pagestyle{empty}


\begin{abstract}
Humanoid robots can benefit from their similarity to the human shape by learning from humans. When humans teach other humans how to perform actions, they often demonstrate the actions, and the learning human imitates the demonstration to get an idea of how to perform the action.  Being able to mentally transfer from a demonstration seen from a third-person perspective to how it should look from a first-person perspective is fundamental for this ability in humans. As this is a challenging task, it is often simplified for robots by creating demonstrations from the first-person perspective. Creating these demonstrations allows for an easier imitation but requires more effort. Therefore, we introduce a novel diffusion model that enables the robot to learn from the third-person demonstrations directly by learning to generate the first-person perspective from the third-person perspective. The model translates the size and rotations of objects and the environment between the two perspectives. This allows us to utilise the benefits of easy-to-produce third-person demonstrations and easy-to-imitate first-person demonstrations. 
Our approach significantly outperforms other image-to-image models in this task.
\end{abstract}

\section{Introduction}
Imitation Learning or Learning from Demonstrations (LfD) is a learning paradigm often encountered in humans which efficiently transmits knowledge about processes. If alternatively an action is taught verbally by explaining it, a further level of abstraction is needed, which demonstrations can bypass. However, learning from demonstrations necessitates a set of sophisticated abilities. In humans, these abilities develops between the ages three and four \cite{c8}. When learning from a demonstration, correctly perceiving the demonstration is the first step. Then, this perceived action has to be transferred to our own perspective so that we can imitate it \cite{visAndSpace}. One of the necessary abilities for this process is perspective-taking.\\

\begin{figure}
    \centering
    \includegraphics[width = 1\columnwidth]{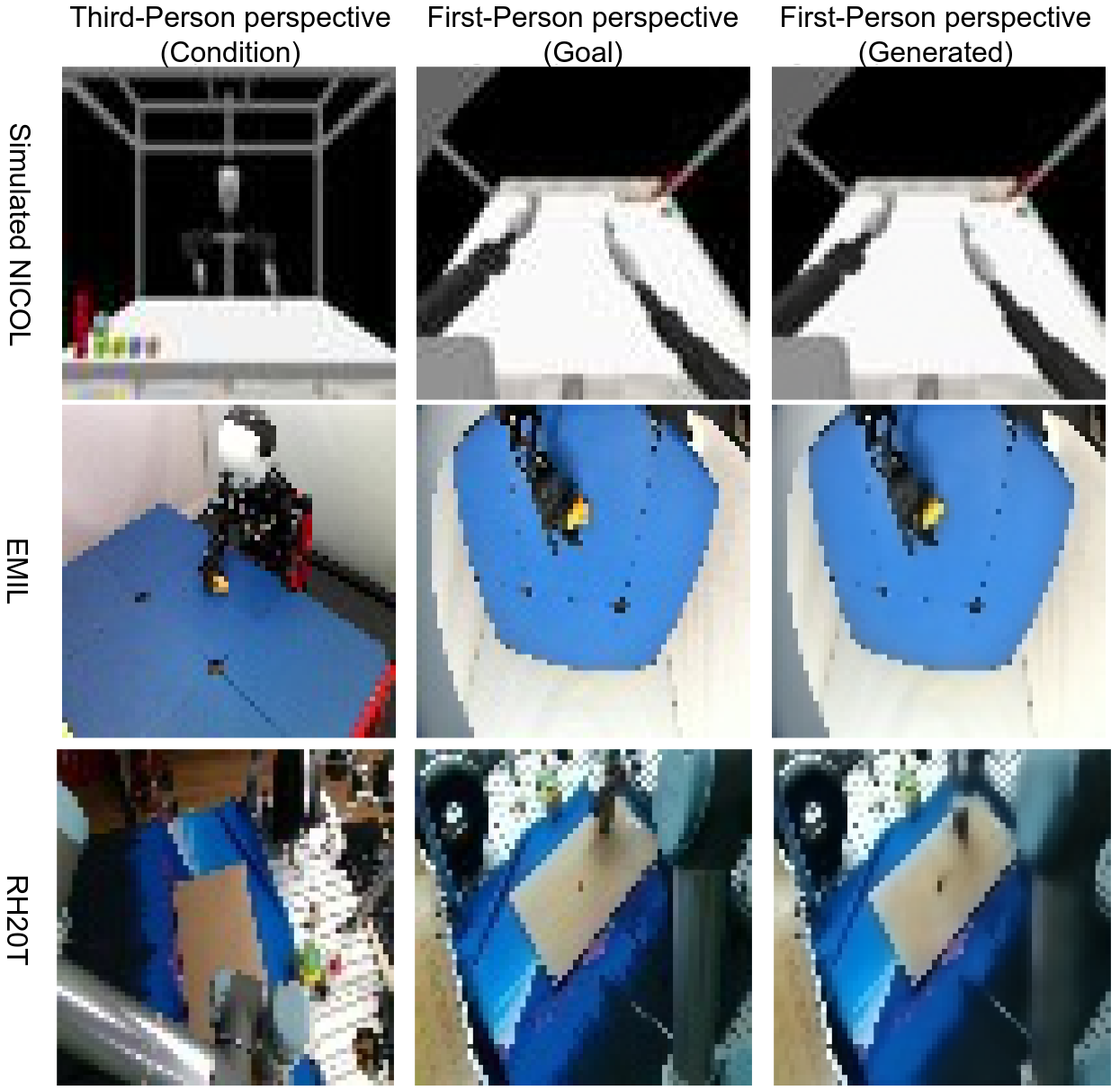}
    \caption{The results of our model for one example of each dataset. Top row is our dataset created with the simulated NICOL robot. An example of the EMIL  dataset \cite{c28} which was recorded in a real-world scenario is in the second row. The third row is for the RH20T dataset \cite{c30} which was recorded in a more cluttered real-world scenario. The resolution is in accordance with the model's input and output.}
    \label{fig:realShowCase}
\end{figure}

Visual and spatial perspective-taking is the ability to see someone and imagine what they are seeing. This ability requires an understanding of the objects that are in someone's field of view and understanding that they will see the objects differently. The objects will have to be transferred from the position, shape, and size as seen by one person to the position, shape, and size as seen by another person.  In this paper, we looked into how a machine learning model can achieve similar abilities. Towards this purpose we developed a diffusion model that utilises images recorded from a third-person perspective to generate corresponding images from a first-person perspective, as shown in Fig. \ref{fig:realShowCase}.\\

The generation of first-person perspectives can be used to generate data for the many existing imitation learning approaches, that learn from first-person perspective demonstrations. This will remove a challenge described by Stadie et al.: \textit{``Unfortunately, hitherto imitation learning methods tend to require that demonstrations are supplied in the first-person [...] While powerful, this kind of imitation learning is limited by the relatively hard problem of collecting first-person demonstrations."}\\

The field of image generation has been quickly improving over the past years through Generative Adversarial Networks (GANs) \cite{c10} and diffusion models \cite{c1,c5,c11}. To generate the first-person perspective, we make use of a diffusion model. Diffusion models work by purposely putting noise on a desired output and then iteratively denoising it again. Through this process, the model learns to generate the desired output from pure noise by detecting and refining general patterns in the desired outputs. The model also learns to build upon its last prediction as the denoising process is iterative, so it also develops the ability to perceive the given images quite well. To direct the generated output, a condition can be added to the diffusion model, such as a label or a prompt. However, diffusion has also been used for tasks in which the condition is what would be the input in classical models, such as an image for object detection \cite{c2}. We similarly have a high-dimensional and complex condition, that should result in a specific generated image. The condition displaying the third-person image is given to the model in various stages to allow it to generate a fitting first-person image.
Our main contributions are:\\

\begin{figure*}[t]
    \centering
    \includegraphics[width=2\columnwidth]{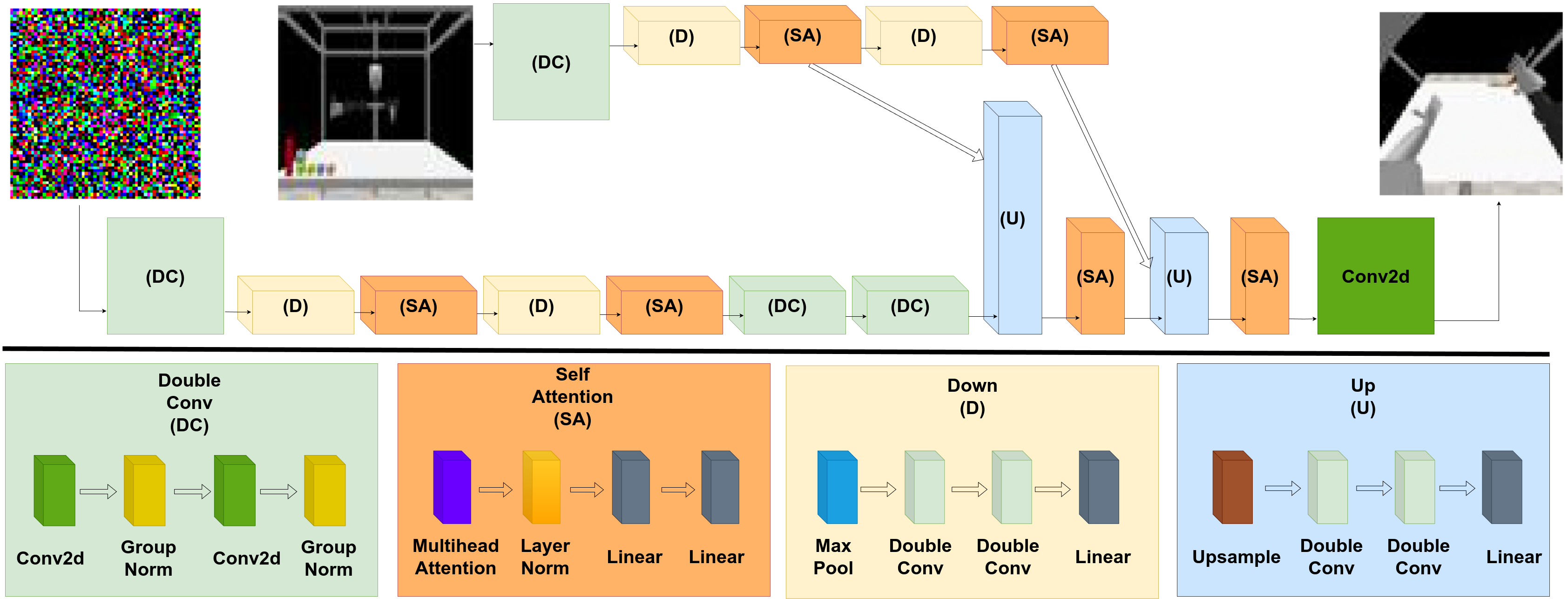}
    \caption{Our architecture has two inputs: the condition, which is the third-person perspective of the robot in the top middle and the noised output during training or, as seen in the top left image, simply noise during inference. Then, the top left image shows the noise that is the input during inference, and the top right image shows the output of our model, the first-person perspective from the robot. Both input images are encoded, each through one of the downward paths. The encoded version of the noise is directly led into the upward path that decodes the image to the output, while the encoded version of the condition is led into the upward path through latent connections.}
    \label{fig:arch}
\end{figure*}

\begin{itemize}
    \item We present a novel architecture to generate images in first-person perspective from images in third-person perspective for a humanoid robot, as shown in Fig. \ref{fig:arch}.
    \item We publish our code and a newly generated dataset with pairs of images for third-person and first-person perspectives.\footnote{We plan to publish our code and dataset at \texttt{https://github.com/knowledgetechnologyuhh}
    \texttt{/robotic-perspective-taking-with-diffusion}}
    \item We evaluate our architecture on three different datasets.
    \item Our model outperforms competing approaches such as pix2pix \cite{c12} and CycleGAN \cite{c15}.
\end{itemize}
\section{Related Work}
\textbf{Generative image-to-image models.} Isola et al. \cite{c12} demonstrate that generative image-to-image models can solve a multitude of tasks. Ranging from changing an image from black and white to coloured, to generating an image just from the edges or turning segmentation labels into a scene image. They achieved this using conditional GANs. Further advancements in image-to-image models have been made by incorporating special losses such as cycle consistency loss, which ensures specific outputs for given inputs \cite{c15}. Going from the third-person perspective to the first-person perspective can also be done using their model, but the results tend to be not as good as for some of the other tasks, as seen in the results section.\\

\textbf{Perspective-Taking.} Specialised models have been developed for this specific task of perspective-taking. To focus the model on a general understanding of the demonstration, the kernel sizes of the convolutional layers were increased, and self-attentive layers were added to improve the understanding of the relations between different parts of the input images. As a demonstration is often a continuous video, past frames can be used to improve the performance \cite{c13}. In a more general approach towards perspective-taking without the direct relation to robots and imitation learning, Liu et al. proposed the Parallel GAN (P-GAN) architecture \cite{c14}. By doing both a transfer from first-person to third-person and, at the same time, a transfer from third-person to first-person, they were able to outperform other models in this task, such as X-Seq or X-Fork \cite{c25} where the GANs were not working in parallel.\\

\textbf{Diffusion Models.} In base diffusion models, a neural network architecture is used that starts with an image input, encodes it and decodes it again to reconstruct it. The key is that different levels of noise are put on the input image. The model, therefore, has to learn to either predict which parts of the image are noise, or what it should look like without the noise \cite{c1}. Once trained, the model can iteratively generate images by progressively removing noise levels encountered during training. As diffusion models possess a purpose similiar to GANs in generating images, it does not come as a surprise that they can be used for many similar tasks, including image-to-image translations. Pix2Pix-zero is one such model, which is able to do zero-shot image translation by changing an image along an ``edit direction" \cite{c15}. Other diffusion approaches focus more on generating images from text prompts, layouts or semantic synthesis \cite{c27}.\\

\textbf{Conditional Diffusion Models.} Similar to GANs, diffusion models can also be conditioned to generate specified output. In diffusion models, the time step is inherently a part of the conditioning process. This parameter specifies the applied level of noise, allowing the network to correctly interpret the input in relation to this noise level. Additional conditions, such as a text embedding to direct the output, can be added to this time step \cite{c16}. The conditioning has also been applied to other tasks apart from image generation, such as action segmentation \cite{c3} or object detection \cite{c2}. In such cases, the output is the bounding box or frame label for the condition input. These works utilise an encoder-decoder structure, in which the condition is added to the input of the decoder layers. In contrast, U-Nets \cite{c4}, which have additional skip connections, are commonly used for image generation tasks \cite{c16} because their skip connections can retain more spatial information.\\

\textbf{Robotic Imitation.} Learning from demonstration is utilised to get past many challenges of machine learning. From a demonstration, much information about how to solve a task can be gained, and the demonstrated solution can be used to recreate that success. Imitation learning can be used to allow robots to learn movement planning. As going from RBG images to joints can be difficult, different methods to record the demonstrations have been used, such as voxels \cite{c20} or point clouds \cite{c21}. Although, for human data, pose detectors exist that are able to find key points of the poses directly from RGB images \cite{c22}, demonstrations are often performed from a first-person perspective so that trajectories stay similar and the demonstration is closer to the imitation. To directly learn from a third-person perspective, the trajectories have to be translated to the first-person perspective, which can be achieved with the use of generative models \cite{c9}. Even just mirroring a pose seen in a mirror can be very challenging for a robot. Hart and Scassellati \cite{c26} develop an approach that makes use of six models: an end-effector model, a perceptual model, a perspective-taking model, a structural model, an appearance model and a functional model for this challenge. Their loss was six times higher for demonstrations from a third-person perspective compared to demonstrations from a first-person perspective. 

\section{Methods}
\begin{figure*}[t]
    \centering
    \includegraphics[width = 1\textwidth]{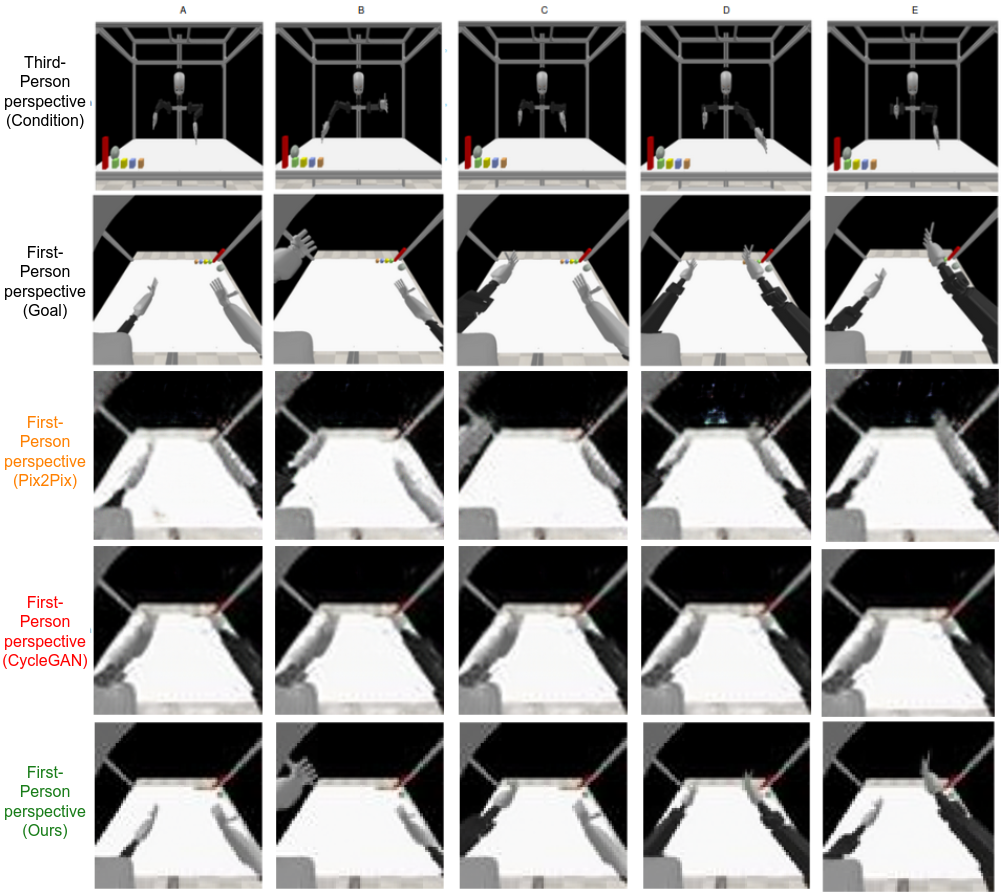}
    \caption{Qualitative comparisons between our model and pix2pix as well as CycleGAN. Five examples (A, B, C, D, E) are shown to illustrate the differences between the models. CycleGan has almost no variance between its output, pix2pix mostly seems to have the correct poses except for example C, but is often unable to correctly recreate the hands and finger structure. Our model has always the correct pose and fully recreates the hands. The resolution is in accordance with the model's input and output.}
    \label{fig:comparisons321}
\end{figure*}
\subsection{Architecture}
Our approach is based on a conditioned diffusion model. In contrast to the model from Ho et al. \cite{c1}, we use a full image as our condition, akin to how DiffusionDet \cite{c2} or  Diffusion Action Segmentation \cite{c3} made use of diffusion models for computer vision tasks. The condition is an image captured from a camera opposite the robot, representing how a person sitting across from it would see the robot. The model then generates the scene as seen by the robot. As the approach uses diffusion, we also have another input, which during inference is simply noise and during training consists of the desired output with noise applied. The level of noise depends on a randomly chosen time step between 0 and 1000 for each sample. A cosine noise schedule calculates the amount of noise depending on the chosen time step \cite{c7}. The cosine schedule allows for a smoother transition from no noise to full noise compared to a linear noise schedule. While the network should mostly rely on the condition, the smoother transition can increase the benefits of the iterative denoising process.\\

The architecture of our model is shown in Fig. \ref{fig:arch}. The structure of the model largely follows a U-Net \cite{c4}, however, there are no latent connections between the primary downward and upward leading paths of the U-Net. Instead, we added a secondary downward leading path that encodes our condition. This path is only connected through the latent connections to the rest of the architecture.
Our model directly predicts the desired output, instead of trying to predict the noise, as proposed by Song et al. \cite{c5}. This decision was made as our goal lies more in correct predictions and accuracy than in the model's ability to generate new ideas or varying outputs, as could be desired when generating images from prompts. During inference, we start with using randomly generated noise for our input and the condition, which is the RBG image taken from a different perspective, providing the information needed to replicate the pose seen in it. Predicting the desired result allows us to utilise only 50 denoising steps, significantly accelerating the inference. Our model leverages both the information provided by the condition and that acquired during later steps of the iterative denoising process to generate its output. \\

Both paths follow the same structure of convolution layers, attention layers and maxpooling layers. The first path ends in two convolution layers that do not exist in the second downward path. The upward path consists of convolutional layers, self-attention layers and upsampling layers in order to get back to the dimensions of the original input. The latent connections from the second downward path to the upward path always go from the self-attention layer to the upsampling layers. We connect the encoding of the condition to the decoder with latent connections so that more spatial information from earlier layers is kept. The time step is encoded and given in the linear layers of the ``Down" and ``Up" modules. \\

The network was trained for 100 epochs. As previously mentioned, the noise applied to each training sample was selected randomly from 1000 possible levels. This extends the variance of our dataset, because the label remains the same regardless of the level of noise applied to the input, and the model should consistently predict this label. The model was trained with the Adam optimiser.

In this end-to-end approach, the network manages to identify the three-dimensional pose of the robot from a two-dimensional RBG image and to transpose that pose from a third-person perspective to an egocentric perspective, which is expressed through a generated RGB image.
\subsection{Datasets}
\textbf{Simulated NICOL.} To create our dataset, we made use of the NICOL simulation introduced by Kerzel et al. \cite{c6}. We use two cameras in the simulation, one situated in the head of the robot and the second one positioned in front of it at a distance that allows us to fully capture the robot. The robot's head is tilted downward to ensure the actions are captured in the first-person perspective. The NICOL robot is a semi-humanoid robot without legs integrated into a collaborative workspace. We make the robot randomly assume poses for both of its arms. In each of the kinematic chains representing the arms, we have 13 joints. Five of these joints are in the hands, while the other eight are in the arms. The joint values range from minus pi to pi, which we scale to be between 0 and 1 for the training. After the poses were assumed, one image from both cameras was captured and the joint values of the robot were recorded. The end-effector position and rotation are also recorded in seven values for each of the hands, with the first three values representing the x, y, and z coordinates and the other four values representing the orientation. During the data collection, 10000 samples were recorded. For training, 80\% of the dataset was used, while 20\% was used for validation. All the images are RBG and have a height and a width of 64 pixels. The random poses come from a uniform distribution but are limited to stay in the visual range and within the ranges of the joints. There are no images where the arms are behind the robot's back or otherwise obscured. The finger joints are always kept in the same position, so only eight of the thirteen recorded joint values per arm are changed.\\

\textbf{EMIL.} The EMIL dataset \cite{c28} was collected with the NICO \cite{c29} robot. During the recording, NICO sat at a table and used its right arm to interact with an object on the table. There are four kinds of interactions: lifting the object, pushing the object, pulling the object or scooting the object to the side. Throughout these interactions, data from NICO and from external sensors was recorded. While data from multiple modalities is recorded, only the visual data from its camera and the external camera are relevant for our approach. The two visual streams are synchronised so that each frame recorded from NICO's camera has a corresponding frame recorded from the external camera. In total, we used 59365 frames from each camera.\\

\textbf{RH20T.} The RH20T dataset \cite{c30} was collected with various robots positioned on a data collection platform. Multiple cameras, microphones, and other sensors collected data while the robots performed various tasks. The background is kept diverse through multiple table covers with complex colour patterns. The RH20T dataset contains demonstrations of more than 140 different tasks. We made use of the frames collected by two of the cameras (cam\_036422060909, cam\_104122063550) that record the robot from different perspectives. We used frames from the RH20T\_cfg3 subset for a total of 187045 frames.

\section{Results}
In this section, we first evaluate our architecture by comparing it with other image-to-image models. The robustness of our model is tested by evaluating it on additional datasets.
Finally, we look at how perspective-taking can improve the generation of joint values.

\subsection{Comparison with pix2pix and CycleGan}

\begin{figure}
    \centering
    \includegraphics[width = 1\columnwidth]{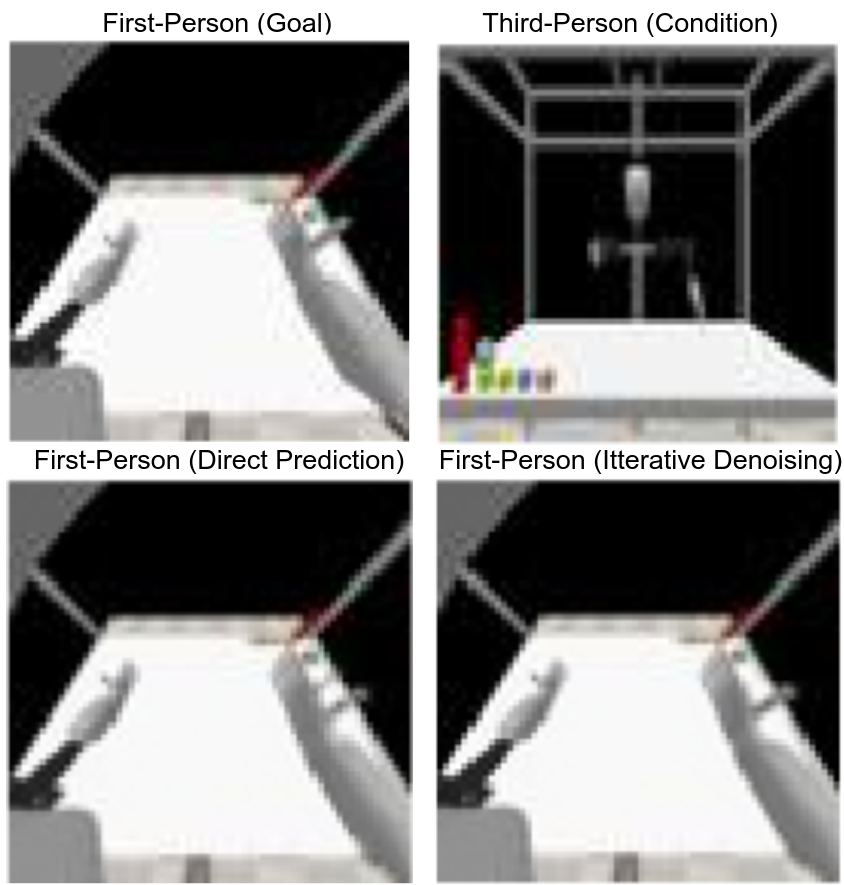}
   \caption{One example of predicting the first-person perspective, where the direct prediction differs from the prediction after the iterative denoising. In the direct prediction, two thumbs can be identified on the hand, whereas only one is left after the iterative denoising, although in the wrong position. The resolution is in accordance with the model's input and output.}
   \label{fig:Error Image Prediction}
\end{figure}


We compare the image generation part of our approach to the pix2pix \cite{c12} model and the CycleGAN \cite{c15} model. We use the implementation provided by the authors of these papers\footnote{https://github.com/junyanz/pytorch-CycleGAN-and-pix2pix/tree/master} and use the standard configurations for both models to train them on our dataset. We evaluate both of these models as well as our own on three metrics with our test set. The metrics used are mean square error (MSE), L1 norm (L1) and the structural similarity index measure (SSIM) \cite{c24}. MSE and L1 simply measure the difference in pixels, while the SSIM uses luminance, contrast and structural comparisons. For MSE and L1, smaller is better, while the opposite is true for the SSIM. Our model outperformed both comparison models across all three metrics, as shown in Tab. \ref{tab:image generation results}.\\

\begin{table}
    \centering
    \caption{Evaluation and comparison of our model, pix2pix and CycleGAN.}
    \begin{tabular}{c|c|c|c}
         Model & MSE $\downarrow$& L1 $\downarrow$ & SSIM $\uparrow$\\
         \hline
         CycleGAN & 0.0221 & 0.0815 & 0.6482 \\
         pix2pix & 0.0252 & 0.0692 & 0.7134 \\
         Ours & \bf0.0007 & \bf0.0086 & \bf0.9773 \\
    \end{tabular}
    \label{tab:image generation results}
\end{table}
Qualitative analysis shows that CycleGAN is unable to correctly adjust to the condition, consistently generating the same image for the first-person perspective, as illustrated in Fig. \ref{fig:comparisons321}. Pix2pix goes one step further with the generated image showing an understanding of the translation from third- to first-person. Compared to our result, however, the hands are often impossible to make out; there are still significant differences in the position and size of the arms and in some cases, artefacts appear in the image. Our results only seem to differ in the resolution compared to the goal images taken directly from the simulation. The hands and arms are in the correct positions, with even the fingers being clearly visible and discernible.\\

Only utilising a single iterative denoising step led to the same MSE of 0.0007, L1 of 0.0086 and a very similar SSIM of 0.9793 showing that the direct prediction of our model can compete with the iteratively denoised prediction.
While the quantitative results showed next to no differences from the iterative denoising, we did come across a few examples where the results changed through iterative denoising. In Fig. \ref{fig:Error Image Prediction}, one example is shown, in which finding the right position of the thumb was difficult for the model, resulting in two thumbs for the direct prediction, which was reduced to one thumb after the denoising process although in the wrong position. There are also cases where some of the blur from the goal images captured in the simulation was smoothed out in the predictions.
\subsection{Comparison with EMIL and RH20T}
\begin{figure}
    \centering
    \includegraphics[width=1\columnwidth]{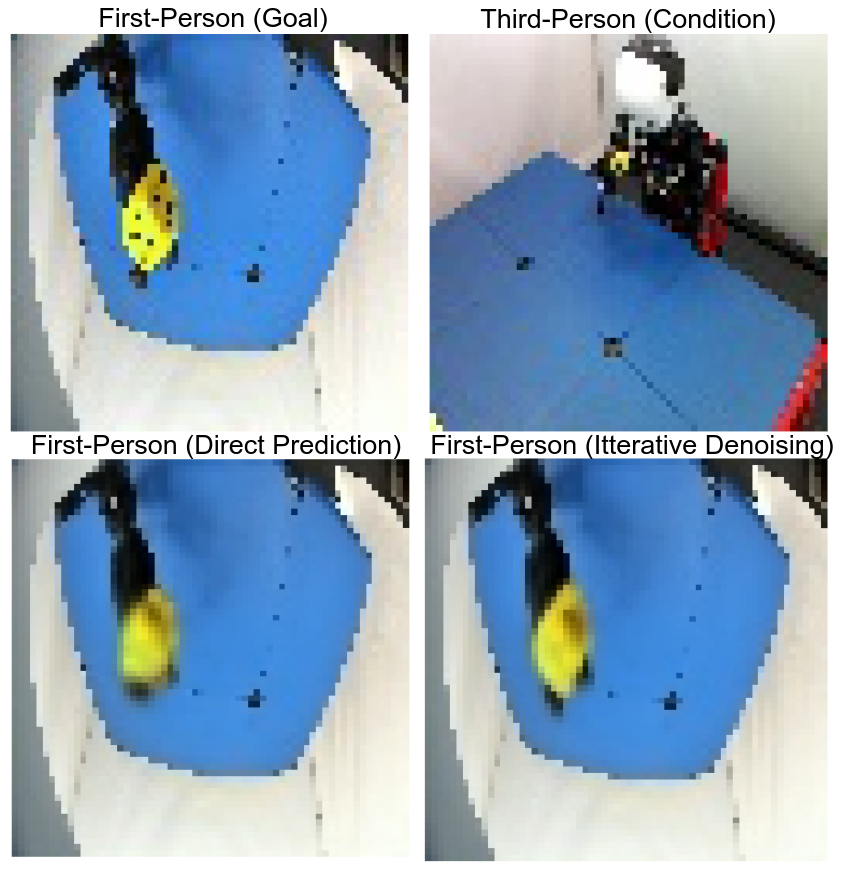}
    \caption{Third-person perspective generation for the EMIL dataset. The resolution is in accordance with the model's input and output.}
    \label{fig:EMILComp}
\end{figure}
\begin{figure}
    \centering
    \includegraphics[width=1\columnwidth]{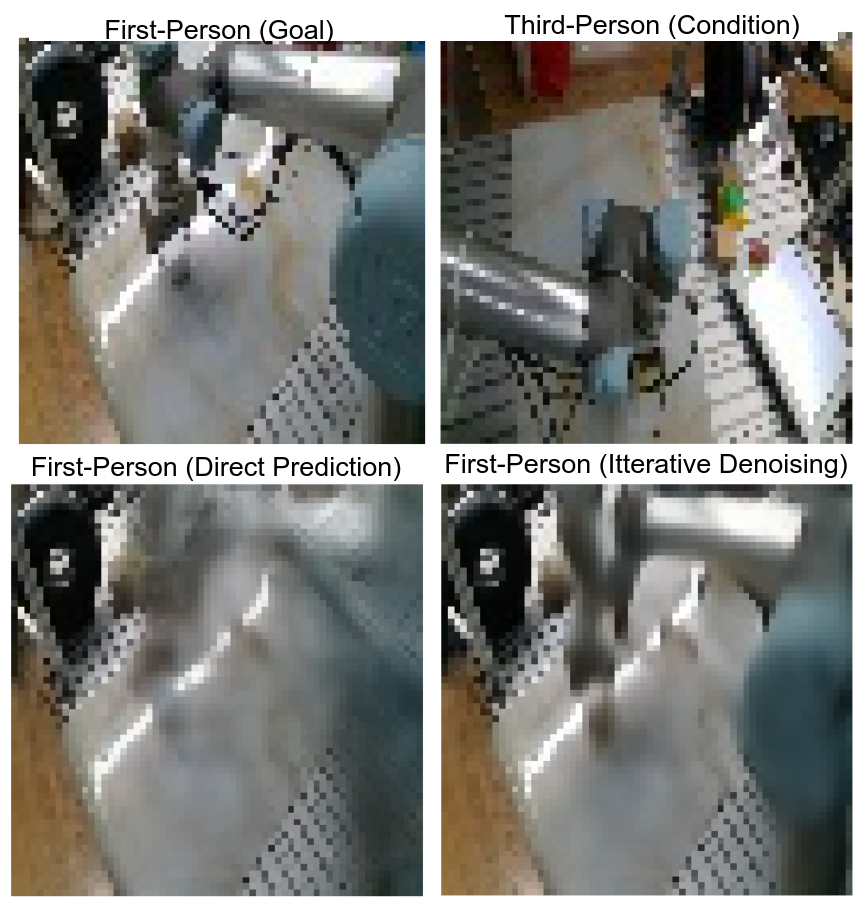}
    \caption{Third-Person perspective generation for the RH20T dataset. The resolution is in accordance with the model's input and output.}
    \label{fig:RH20TComp}
\end{figure}
\begin{figure*}[t]
 \centering
 \includegraphics[width=2 \columnwidth]{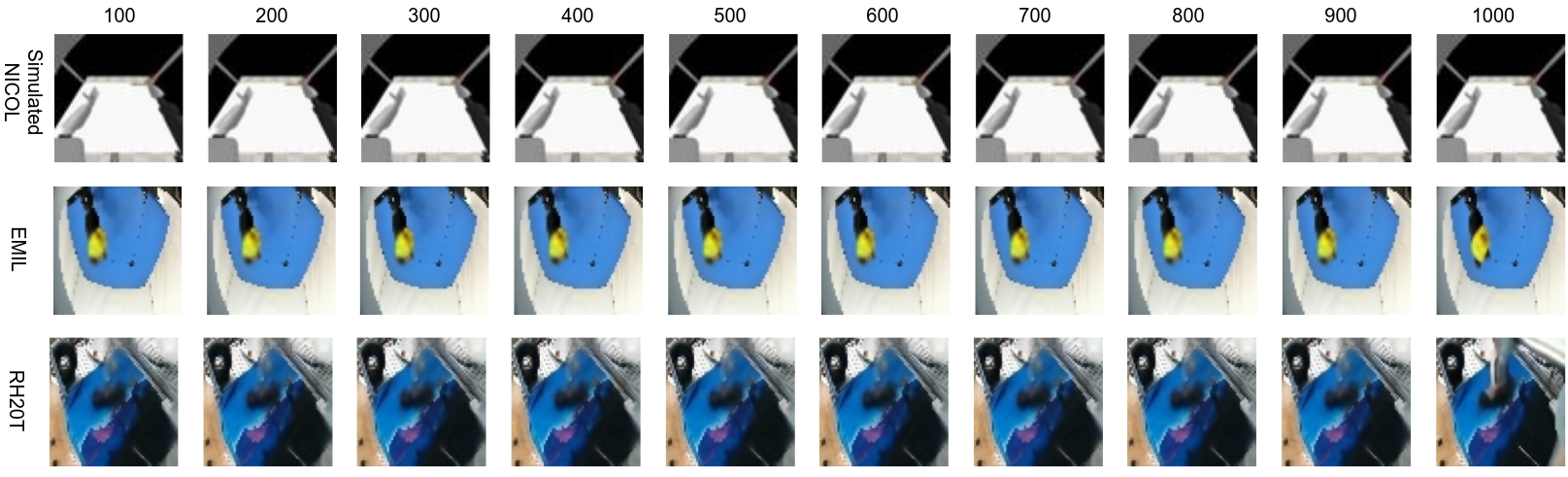}
 \caption{Comparison for the development over the iterative denoising for each of the datasets, shown are the generated images after step 100, 200, 300, 400, 500, 600, 700, 800, 900, 1000 from left to right.}
 \label{fig:timeComp}
\end{figure*}
To further evaluate the model, we test it on two additional datasets: the EMIL dataset \cite{c28} and the RH20T dataset \cite{c30}, again the model was trained on 80\% of these datasets and tested on the remaining 20\%. While our dataset was collected in a simulated environment, both of these datasets were collected in the real world. Compared to the simulated environment, there are far more details in the collected frames of the real world. In the EMIL dataset, the background is mostly kept clean, while the RH20T dataset has various shapes and colour patterns in each frame. The RH20T dataset also has the added challenge of changing table covers, while the only thing that changes in the EMIL dataset is the small object, and no external objects change in the simulated dataset. The datasets have different levels of complexity that might be encountered in the field of imitation learning.\\

For each dataset, we trained a model with the exact same architecture and used the SSIM, MSE and L1 metrics to compare them. We also tested different amounts of iterative denoising steps, as shown in Tab.\ref{tab:dataSetComparison}, to see if the complexity of the dataset had an influence on the behaviour during inference. The results align with our expectations, indicating that the simulated dataset is the least challenging followed by the EMIL dataset and finally the RH20T dataset. Even for the more complex datasets, the model is capable of realistically generating new perspectives. In the EMIL dataset, the difference between the real first-person perspective and the generated first-person perspective is difficult to discern, even without iterative denoising steps, as shown in Fig. \ref{fig:EMILComp}. For the RH20T dataset, the model is also able to generate a new perspective, although small details remain challenging. There is a significant difference between the generation without any iterative denoising steps and the generation after 50 iterative denoising steps, as shown in Fig. \ref{fig:RH20TComp}.\\

The more difficult datasets necessitate iterative denoising to generate finer details and to de-blur parts of the image. Surprisingly, this is not reflected in the quantitative results, likely because the differences only affect small parts of the images. Furthermore, sharper images do not necessarily result in better pixel matchings, as they might be slightly misaligned. To further explore the effect of the iterative denoising, we show one example for each dataset in Fig. \ref{fig:timeComp}, in which a full 1000 iterative denoising steps are used during the inference, of which every one-hundredth step is displayed. Only the last jump in denoising steps shows a change.
For the RH20T dataset, the robot arm becomes much clearer, with the right-most part only appearing during the last step. In the EMIL dataset, the last step makes the fingertips of NICO's right hand clearer.

\begin{figure}[t!]
    \centering
    \includegraphics[width = 1\columnwidth]{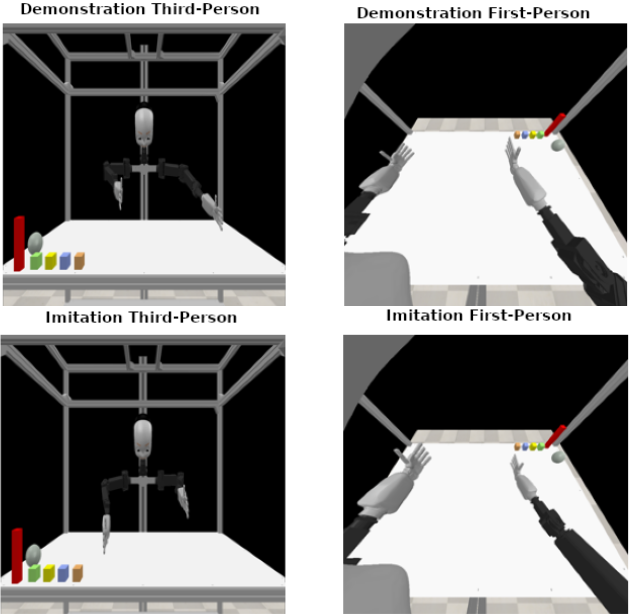}
    \caption{One example of predicting joint values from the first-person perspective, where the imitation looks similar in the first-person perspective but different in the third-person perspective compared to the demonstration.}
    \label{fig:jointPrediction}
\end{figure}

\begin{table}
    \centering
    \caption{Comparison between datasets and amount of iterative denoising steps. Errors are consistent across datasets, with higher errors for the more complex datasets.}
    \begin{tabular}{c|c|c|c}
         Model & MSE $\downarrow$ & L1 $\downarrow$ & SSIM $\uparrow$ \\
         \hline
         Simulated (1 steps) & 0.0007& 0.0086 & 0.9793 \\
         EMIL (1 steps& 0.0014 & 0.0183 & 0.9151\\
         RH20T(1 steps & 0.0101& 0.0603 & 0.7606\\
         \hline
         Simulated (200 steps) & 0.0007 & 0.0083 & 0.9805\\
         EMIL (200 steps& 0.0015& 0.0181 & 0.9156 \\
         RH20T(200 steps & 0.0133 & 0.0610 & 0.7628 \\
    \end{tabular}
    \label{tab:dataSetComparison}
\end{table}

\subsection{Joint Generation}

To assert the benefit of the generated first-person perspective, we adjust our architecture by changing the last layer to output joint values instead of RGB images. Then, we train the adjusted architecture with the third-person images once and once with the generated first-person images. We use the simulated NICOL dataset for this purpose. Using 20\% of the dataset for validation, the model achieves a mean squared error of 0.0027 for the joint values with the third-person images. It appears that the model finds it challenging to generalise from the training set, where we are able to get to a loss as low as 3.3e-6. While the general configuration seems to go in the correct direction, often, some of the joints are not completely accurate, shifting the whole arm in a wrong direction.
Training on the generated first-person images, instead we managed to reduce the training error to 3e-7 and the validation error to 0.0017, clearly showing the benefit of the first-person perspective. One example of the results from using first-person images for the condition is shown in Fig. \ref{fig:jointPrediction}. \\

It seems to be more difficult to directly predict the joints instead of recreating the first-person image, despite both requiring a spatial understanding through the third-person image. One possible reason for that is the redundancy in the arm \cite{c17,c18,c23}. Allowing for many poses that look similar while having very different joint configurations. Fig. \ref{fig:jointPrediction} shows one particular interesting result demonstrating this, where the first-person perspective looks very similar, while the actual pose looks quite different. With so many very similar-looking poses, the model tends to predict values closer to the mean. 
\section{Conclusion}
We observed a difference between quantitative analysis and qualitative analysis regarding the effect of iterative denoising steps. The quantitative analysis did not show clear improvements from using a higher number of noise steps, while the qualitative reveals a clear effect, especially the later denoising steps. The benefit of the later denoising steps was particularly high for the more complex datasets, showing how diffusion models have unique qualities that can be helpful when handling complex data. The difference between the quantitative and qualitative results also highlights a need for further image comparison metrics that can evaluate higher levels of abstractions beyond pixel-wise similarities.\\

The benefits of generating first-person images were shown through their use in generating joint values. A significant difference in performance was observed between using third-person perspective images and first-person perspective images. Based on this, we assume that this way of generating first-person perspectives can be useful for other approaches that work on imitation, be it as an additional modality to an existing third-person perspective or as a way of generating more training data. Since higher amounts of data did not have a positive effect on the results, we assume that having a noisy version of the desired output available during training can speed up the time it takes the model to learn consistent patterns across the datasets.\\

Throughout this paper, we have demonstrated that our approach effectively learns spatial relations and can resize and rotate objects to adjust them to a new perspective. This enables our approach to generate first-person perspective images from third-person perspective images, that are nearly indistinguishable from actual first-person perspective images, and surpass established approaches such as pix2pix or CycleGAN. The architecture has shown promising results across three different datasets, each with unique properties. 

\end{document}